# Brain-inspired global-local learning incorporated with neuromorphic computing

Yujie Wu[1], Rong Zhao[1], Jun Zhu[2], Feng Chen[3], Mingkun Xu[1], Guoqi Li[1], Sen Song[4], Lei Deng[5], Guanrui Wang[1], Hao Zheng[1], Songchen Ma[1], Jing Pei[1], Youhui Zhang[2], Mingguo Zhao[3], and Luping Shi[1*]

[1]Center for Brain-Inspired Computing Research (CBICR), Beijing Innovation Center for Future Chip, Optical Memory National Engineering Research Center, Department of Precision Instrument, Tsinghua University, Beijing, China

[2]Department of Computer Science and Technology, Tsinghua University, Beijing 100084, China

[3]Department of Automation, Tsinghua University, Beijing 100084, China

[4]Laboratory of Brain and Intelligence, Department of Biomedical Engineering, IDG/ McGovern Institute for Brain Research, CBICR, Tsinghua University, Beijing, China

[5]Department of Electrical and Computer Engineering, University of California, Santa Barbara, CA 93106, USA

*e-mail: lpshi@mail.tsinghua.edu.cn.

# Abstract

Two main routes of learning methods exist at present including error-driven global learning and neuroscience-oriented local learning. Integrating them into one network may provide complementary learning capabilities for versatile learning scenarios. At the same time, neuromorphic computing holds great promise, but still needs plenty of useful algorithms and algorithm-hardware co-designs for exploiting the advantages. Here, we report a neuromorphic hybrid learning model by introducing a brain-inspired meta-learning paradigm and a differentiable spiking model incorporating neuronal dynamics and synaptic plasticity. It can meta-learn local plasticity and receive top-down supervision information for multiscale synergic learning. We demonstrate the advantages of this model in multiple different tasks, including few-shot learning, continual learning, and fault-tolerance learning in neuromorphic vision sensors. It achieves significantly higher performance than single-learning methods, and shows promise in empowering neuromorphic applications revolution. We further implemented the hybrid model in the Tianjic neuromorphic platform by exploiting algorithm-hardware co-designs and proved that the model can fully utilize neuromorphic many-core architecture to develop hybrid computation paradigm.

# Main text

Realizing human-like learning has been a long-term goal that has been sought through various means of artificial intelligence (AI). Error-driven global learning represented by the backpropagation family, can learn sophisticated mappings from abundant data and even surpass the human-level performance in some specific tasks, such as processing of images[1] and playing games[2]. Another main route is neuroscience-oriented correlation-driven local learning, which mainly focuses on exploring advanced brain-inspired learning mechanisms with remarkable computational efficiency[3-5], which are represented as the Hebbian and spike timing-dependent learning (STDP).

Both approaches are built on neural networks, have specific advantages and share the same vision to advance intelligence with learning capability[6-8]. However, neither method currently outperforms the other on all problems, and both methods are still inferior to human-level performance on many general tasks. Thus, it is considerable to leverage their different advantages and integrate them into one single hybrid model for exploring synergic learning in multiple general tasks. At the same time, by mimicking the way biological spiking neurons work, the neuromorphic computing provides a great substrate for modelling local learning with high efficiency. It is one of the two most promising technologies recommended by the Semiconductor Industry Roadmap to further improve computation efficiency beyond Moore's law[9]. A recent review paper[5] in *Nature* shares the perspective that '*neuromorphic computing as an energy-efficient way to enable machine intelligence through synergistic advancements in both hardware (computing) and algorithms (intelligence)*'. More recently, Tianjic, a unified neuromorphic computing platform has been developed, which provides a hybrid hardware platform supporting various neuroscience-oriented and computer-science-oriented models with high efficiency[8]. Hence, it is highly expected to develop the hybrid learning algorithms with algorithm-hardware co-designs in neuromorphic platform to advance both learning capability and efficiency. However, because of the fundamental differences in their learning characteristics and complex spiking dynamics, developing such hybrid learning model in neuromorphic platforms is challenging.

In general, backpropagation and most of its variants yield two alternative information circuits (top-down and bottom-up) and communicate with continuous signals. Its powerful learning capability largely benefits from hierarchical feature abstractions via layer-by-layer allocation of global supervision errors. In contrast, neuroscience-oriented local learning prominently occurs between presynaptic and postsynaptic neurons and is triggered by asynchronous event-driven spike activity. (Note that such correlation-driven weight updates computed over adjacent neuronal activity are called 'local plasticity' (LP) in the context and the former are called 'global plasticity' (GP) for distinguish). These two learning approaches exhibit independent learning circuits and weight update behaviours. Furthermore, when incorporating discrete spike representation and complex and diverse neuron dynamics, it poses great challenges to the development of hybrid learning on neuromorphic models.

Currently, hybrid model is a frontier topic of broad and diverse interests. One main thread exploits the three-factor rules[10] to combine these two learning methods and develop more biologically plausible learning models, such as temporal difference learning, rewarded STDP, and random backpropagation[11-19]. For example, some works took global errors as the third factor to modulate the magnitude of local weight update and introduced random feedback connections for the propagation of error signals, thereby achieving a fully local computation[11-13]. Some works explored the three-factor learning rules in a spiking probabilistic framework to approximate bio-plausible inference and learning solutions[14-17]. Some works utilized the three-factor rules to seek the biological interpretations of backpropagation[18,19]. With a different purpose, another thread targets to utilize diverse characteristics of these two learning methods so as to solve many complicated problems[20-27]. One approach modelled the local modules as augmented modules and optimized the connection weights of local modules to improve network performances on image classification task[20] or sequential tasks[21].

Alternatively, another closely-related approach used meta-learning methods to optimize the local learning rules. Meta-learning generally refers to the algorithms to improve or learn the model's learning ability. In the early 1990s, ref.[22] proposed to use global supervision signals to optimize the parameters of local learning rules. In the follow-up work, several works extended this framework to optimize the parameters of local plasticity by gradient descents and established large-scale non-spiking networks for solving few-shot learning[23,24], reinforcement learning[24,25] or unsupervised learning[26,27]. Despite of recent tremendous progresses on meta-learning-based models, few studies considered neuromorphic computing and practical on-chip learning. Owning to the various differences between the two learning methods, it still lacks an effective model that can integrate these characteristics to explore the advantages of hybrid learning in multiple complex learning scenarios, such as fault-tolerance learning and continual learning, and support efficient hardware realizations. In addition, spiking neural networks (SNNs), a representative neuromorphic model with the unique features of spatio-temporal dynamics and rich coding schemes, are naturally suited for modelling temporal-based and rate-based local learning and enabling event-driven low-power computation. Currently, most of SNNs are established using either single global learning or non-learnable local plasticity[28-38]. A general method that can learn to optimize spike-based local plasticity and integrate the complementary advantages of global-local learning in neuromorphic applications and neuromorphic hardware has yet to be explored.

Here we report a spike-based hybrid plasticity (HP) model based on a brain-inspired meta-learning circuit and the bi-level optimization technique[39,40], and explore its advantages on a variety of new learning scenarios and practical implementations on dedicated neuromorphic chips. By jointly deriving from ion-channel dynamics and membrane potential dynamics, we integrate these two learning approaches into a differentiable leaky and integrated fired model with short-term synaptic plasticity. In this manner, we demonstrate that with small modifications of LP modules, the hybrid model can solve three different learning problems, including few-shot learning, continual learning, and fault-tolerance learning, and we further analyse the model effectiveness. Finally, we exploit the method of the algorithm-hardware co-design by designing a new mapping scheme with developed cycle-accurate hardware simulator, and evaluate the efficiency on the Tianjic neuromorphic platform[8].

# 1 Spike-based hybrid plasticity model

We develop a spike-based hybrid learning model based on a general framework of the biological three-factor learning rule[10]. In particular, our design principles further exploit the two experimental cues from neuroscience[41-46] regarding versatile synaptic modulation behaviours and biological multiscale learning mechanism to integrate these two learning circuits in a meta-learning paradigm (Fig. 1a, b). The differences from the previous three-factor learning rules are elaborated in the *Discussion*.

(1) Numerous *in vitro* studies have showed that in the hippocampus, local neural circuits can be controlled by various top-down neuromodulators, such as dopamine and acetylcholine, also known as by the third factor, which inform the local circuits about behavioural performance[10]. These modulating factors arrive at many synapses in parallel and can modulate various plasticity behaviors, such as the update rate and the plasticity consolidation[10,41,42]. It indicates that the versatile local plasticity behaviors can be controlled by the top-down signals and these factors can be formulized as meta-learning parameters on synaptic plasticity in a weight-sharing manner;

(2) These neuromodulators allow to encode, for examples, rewards and supervision errors, on multiple temporal scales. The evolution scales and learning manners of these neuromodulators are radically different from synapses and occur on multiple temporal scales[42-44]. This coupling of plasticity and neuromodulators is an important mechanism of brain in building behavior functions[42,45], such as the regulation of muscle contraction[45]. It further implies a multiscale learning mechanism that the learning process of neuromodulators and synaptic weights may happen in different scales and thus they can be formulized as two types of variables with different learning manners in optimization process.

On this basis, we first formulize the hyperparameters of local plasticity, such as the learning rate and sliding threshold, as a group of meta-parameters $\boldsymbol{\theta}$. Furthermore, since these meta-parameters $\boldsymbol{\theta}$ are to control weight update behaviours, we model the $\boldsymbol{\theta}$ as the upper-level variable of connection weights $\boldsymbol{w}$ and transform the hybrid learning process into a bi-level optimization problem[39,40] as in refs.[39,40,47,48]. It allows us to decouple the optimization process of $\boldsymbol{\theta}$ and $\boldsymbol{w}$ and use the hyper-parameter optimization technique[39,40] to optimize $\boldsymbol{\theta}$ (see *Methods*). In this manner, we expect to provide an effective and flexible optimization strategy that the local module can efficiently learn the learning rule and produce useful augmented signals of the correlation-based input patterns to facilitate neuron computations.

At the same time, we incorporate the complex spiking dynamics, local plasticity and global learning into one unified temporal credit assignment problem. To this end, we firstly derived from membrane potential dynamics and ion-channel dynamics and obtained a differentiable signal propagation equation of spiking neurons (see *Methods*). Then, we used the backpropagation through time (BPTT) algorithm[49] as global learning to train the spiking model and adopted the surrogate functions[37,38] to approximate the derivative of spiking firing function. Because local plasticity has an independent correlation-based updating manner, direct configuration of local modules with handcraft parameters is difficult to ensure the convergence of hybrid models. Thus, it is worth considering to incorporate the impact of local weight updates into the entire optimization framework, thereby integrating it into the global learning. As illustrated in Fig. 1c, we observe that *local plasticity is mainly affected by the adjacent neuron activations in the current state, and in turn affects the spread of spike signals in time and space*, and thus take an equivalent parametric modelling strategy that can transform the local synaptic increments $\Delta w$ into a temporal parametric function related to presynaptic spike activity (e.g. the pre-synaptic spike $x_1^t$ in Fig. 1c), postsynaptic spike activity (e.g. the post-synaptic spike $s_1^t$ in Fig. 1c) and some local hyperparameters. By doing so, we exclusively model local weight updated behaviours as a temporal-based function with respect to adjacent neuron activity and meta-parameters. This ensures that the local plasticity can be transformed into a parametric function and the effect of synaptic modifications can be further incorporated in the temporal credit assignment problems. Interestingly, we note that this transformation is coincident with a variant of synaptic dynamics. More specifically, derived from the synaptic differential dynamics, the synaptic weights $w(t)$ have two terms

$$w(t) = w(t_n)e^{\frac{t_n-t}{\tau_w}} + \tilde{P}(t, pre, post; \boldsymbol{\theta}) \triangleq w_{GP} + w_{LP}, \quad (1)$$

where $w(t_n)$ denotes the phase value of synaptic dynamics at discrete time $t_n$, $k(t) = e^{\frac{t_n-t}{\tau_w}}$ denotes the synaptic decay function, $\tau_w$ denotes the synaptic constant. Here $\tilde{P}(t, pre, post; \boldsymbol{\theta})$ denotes generic local modifications controlled by the presynaptic activity, $pre$, the postsynaptic activity, $post$, and the local layer-wise meta-parameters $\boldsymbol{\theta}$, consisting of local learning rate, sliding threshold and some other hyper-parameters determined by specific local learning rules we used. On this basis, if we further assume the top-down signals to modify the state $w(t_n)$ at the specific time $t_n$ when supervision signals are provided, and assume $\tilde{P}(t, pre, post; \boldsymbol{\theta})$ to represent local plasticity driven by neuron activities and meta-parameters $\boldsymbol{\theta}$, then equation (1) can be used to accordingly decompose the weight into two parts, $w_{GP}$ and $w_{LP}$.

In this manner, we develop a global-local hybrid learning model (Fig. 1d), which enables a differentiable spiking model incorporating diverse neuronal dynamics behaviours and can reconcile the different features of LP and GP learning by allocating them to act on different weight fractions and time scales. In the inference phase, the network receives input stimulus and propagates the signals in an event-driven manner. In parallel, the local plasticity is driven by the neuronal co-activity in the adjacent layers. In the supervised phase, the errors of supervision act as instantaneous feedback signals to adjust the synaptic value and are also used to optimise the local meta-parameters.

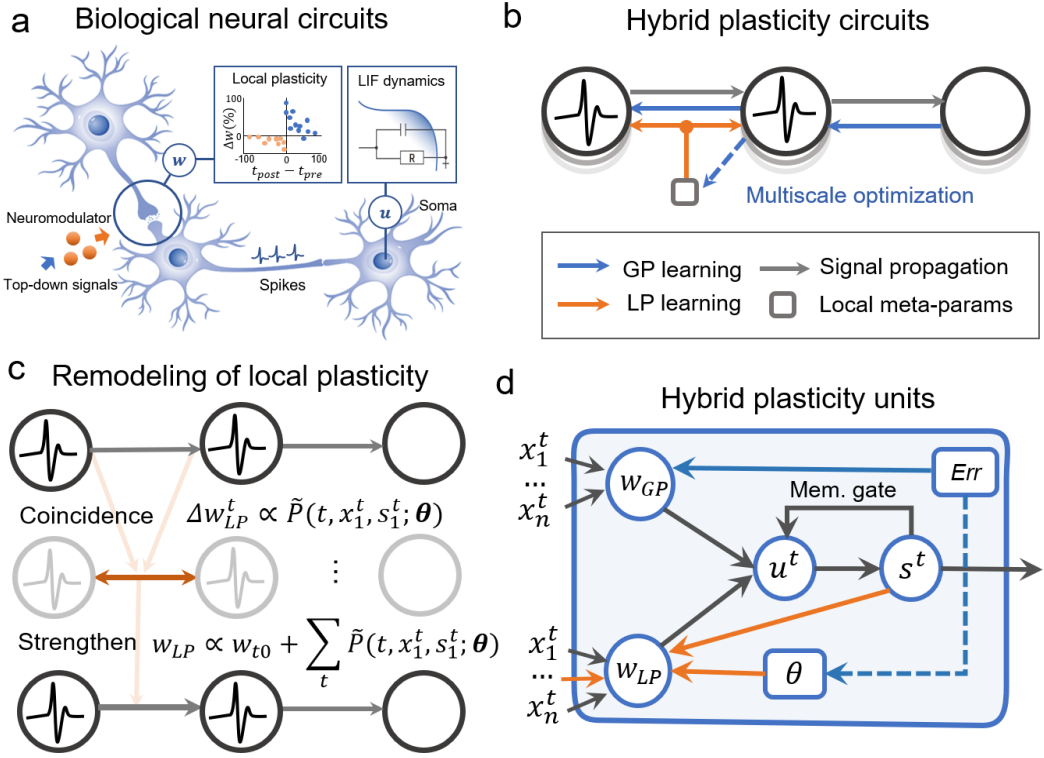

**Figure 1. Illustration of hybrid plasticity (HP) model**. **a,** An illustration of biological synaptic plasticity mechanisms and neuronal dynamics. The neuromodulators can encode top-down supervision information into local circuits resulting in sub-optimal behavioural performance. They exhibit a radically different evolution (learning) process from synaptic plasticity. **b,** motivated by the neural circuitry, this work develops a multiscale meta-learning paradigm to integrate these two types of learning in a unified neuromorphic model. It models the parameters of local learning as a type of meta-learning parameters and decouples the learning process of weights (blue solid lines) and these local meta-parameters (blue dash lines) by using bi-level optimization technique, enabling a flexible multiscale learning. **c,** we remodel the biological short-term plasticity as a parametric function of neuron activity over time. Taking Hebbian learning as an example. Local weight modifications can be equivalently modelled as a parametric function related to the weight initial value $w_{t_0}$, the presynaptic spike activity $x_1^t$, the postsynaptic spike activity $s_1^t$, and the hyperparameters of Hebbian rule (i.e., learning rate and weight decay). **d,** the basic HP unit divides the weight into two parts, $w_{GP}$ and $w_{LP}$. $w_{GP}$ is updated based on global supervision errors, and $w_{LP}$ is updated based on adjacent neuron activities and meta-parameter parameters. In each HP unit, we use a memory gate that controls the reset or leaky integration behaviours of membrane potential $u_t$ by using spike firing state $s_t$.

## 2 Results

### 2.1 Comprehensive performance evaluation

Firstly, we comprehensively evaluated the basic performance of the HP models on the
9

accuracy, coding efficiency and convergence using image classification datasets (including MNIST, Fashion-MNIST and CIFAR10) and neuromorphic datasets (including CIFAR10-DVS and DVS-Gesture). The network architecture and training details are described in the *Methods*.

Table 1: Comparison of the state-of-the-art results of spike-based networks on several image classification tasks and sequential-based classification tasks. The results of the proposed models in this work collected from ten repeated runs and ± indicates the 95% confidence interval.

| Model | Dataset | Method | Avg. latency.[b] | Accuracy (%) |
|---|---|---|---|---|
| Spiking MLP[28] | MNIST | LP | 350 | 95.00 |
| Spiking MLP[29] | MNIST | GP | - | 97.55 |
| Spiking CNN[30] | MNIST | GP Converted.[a] | 200 | 99.12 |
| Spiking CNN **(This work)** | MNIST | **HP** | 7.5 | 99.50±0.04 |
| Spiking MLP[31] | F-MNIST | GP based | 400 | 90.13 |
| Spiking CNN **(This work)** | F-MNIST | **HP** | 3.9 | 93.29±0.07 |
| Spiking CNN[32] | CIFAR10 | GP Converted.[a] | 500 | 91.55 |
| Spiking CNN[33] | CIFAR10 | GP Converted.[a] | 5 | 91.78 |
| Spiking CNN **(This work)** | CIFAR10 | **HP** | 4.5 | 91.08±0.09 |
| Spiking CNN[38] | CIFAR10-DVS | GP | 50 | 60.05 |
| Spiking CNN **(This work)** | CIFAR10-DVS | GP | 50 | 67.22±0.43 |
| **Spiking CNN (This work)** | **CIFAR10-DVS** | **HP** | 50 | 67.81±0.34 |
| Spiking CNN[34] | DVS-Gesture | GP | 400 ms | 94.13 |
| Spiking CNN[35] | DVS-Gesture | GP | - | 96.53 |
| Spiking CNN[36] | DVS-Gesture | GP | 400 ms | 96.78 |
| **Spiking CNN (This work)** | DVS-Gesture | GP | 400 ms | 96.21 ± 0.32 |
| **Spiking CNN (This work)** | **DVS-Gesture** | **HP** | 400 ms | 97.01 ± 0.21 |

.[a] converted from a pre-trained GP-based ANN; .[b] Results from the training time steps or windows reported in the published work.

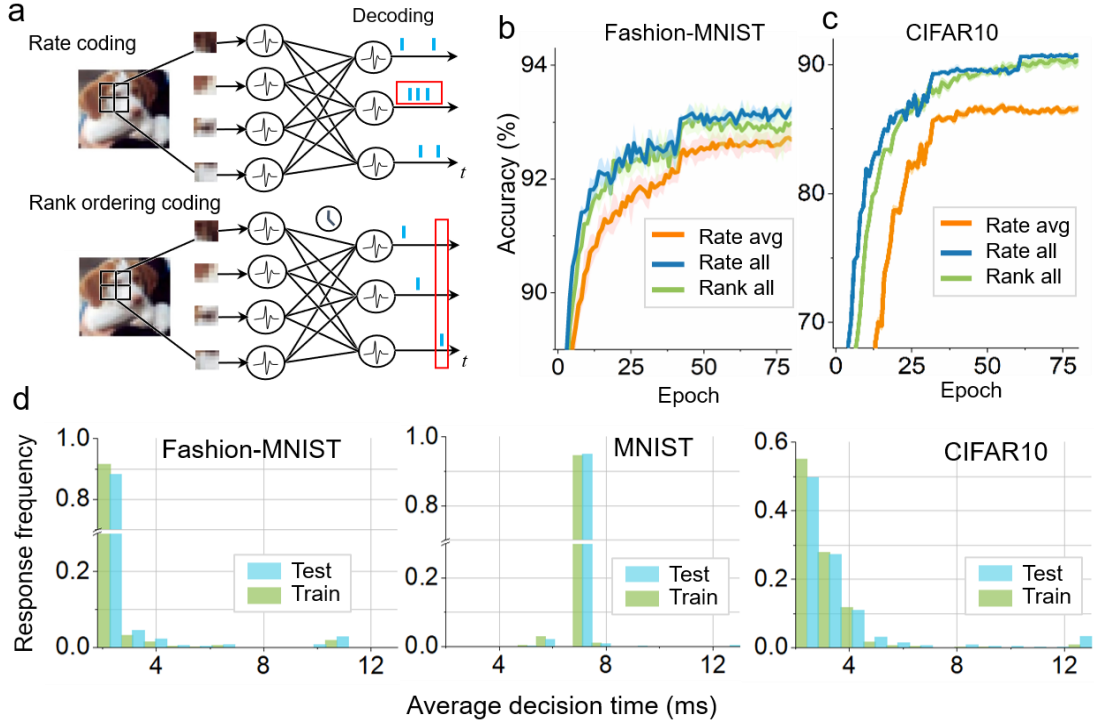

**Figure 2. The hybrid plasticity spiking neural network can flexibly support different coding schemes and achieve a trade-off between performance and efficiency. a,** Illustration of rate coding and temporal rank order coding. In rate coding, the network counts the number of spikes fired by the output neurons as the output representation. In temporal rank order coding, the firing time of each neuron is regarded as encoding of additional information, and the network determines the output in a winner-take-all manner where the firing order of the output neurons is used to output classification results. **b-c,** Comparison of the training curves of the model with different coding schemes (called Rate-all or Rank-all) for the Fashion-MNIST (**b**) and CIFAR10 (**c**). The average and standard derivations over ten repeated trials are plotted. The average firing time over ranking coding schemes is counted and rounded up, and the rate-based model is tested with this average time window (called Rate-avg.). **d**, Average response time for each testing category. We recorded the average response timesteps required to produce results for each testing sample when running SNNs with rank order coding in an event-driven inference mode. After that, we counted these time steps and generated a normalised frequency histogram.

We first analysed its accuracy in different types of datasets in Table 1. Because SNNs achieve a balance between performance and efficiency, here we mainly compared our model with other SNN models. In image classification tasks, the HP SNN with the rank ordering coding[50] (see *Methods*) achieves higher accuracies compared with other published work and significant lower average coding latency than other models. In sequential learning tasks, the HP SNN improves the performance than single GP SNN, which indicates that local modules are beneficial for the use of longer learning time-scales. We give a more in-depth analysis in Fig. 3 thereon.

In recent years, more and more SNN works have begun to leverage rich neuronal dynamics for improving learning capability[11,12,37,51,52]. One prominent feature of our work is the modelling of synaptic ion-channel dynamics, thereby finding an interesting relationship between spiking information coding and an event-driven inference capability (see *Methods*). Next, we show that the HP SNNs can support rate coding and temporal rank coding schemes (see Fig. 2a) and analyse the coding efficiency in image classification task. Fig. 2b-c compare the results of different coding methods in Fashion MNIST and CIFAR10. Specifically, we compared the training curves of the SNNs with rank order coding and the ones with rate coding using the same simulation time windows. Because of the event-driven property of rank order coding, we computed its average time windows and further compared the training curves of SNN with rate coding using the same average time windows (called Rate-avg.) in Fig. 2b-c. It indicates that HP SNNs are well suited for rate-based and rank-based spiking networks. With a longer average time window, the rate-based model obtains higher accuracy. With more flexible and event-driven response characteristics, the rank-based model achieves lower inference latency with a slight accuracy loss. Furthermore, we also plot the details of the average response time of each category for different datasets in Fig. 2d. We found that HP SNNs use a flexible strategy for decision-making. Using the CIFAR10 dataset as an example, for most categories, HP SNNs can make decisions within four time-windows, while for some complicated patterns, the model requires more time to make decisions. This flexible decision process significantly reduces the average inference latency and leverages the high efficiency of neuromorphic hardware. We instantiated our model on the Tianjic chips and reported the energy evaluation in *Supplemental Table S1*.

We further compared the convergence of the proposed approach with other related learning methods. Intuitively, since local learning and global learning have independent update methods, directly combining them cannot ensure convergence. Furthermore, regarding neuroscience-oriented learning models, such as STDP-based SNNs, systematically and generally configuring parameters over local learning rules has yet to be resolved. Although hand-designed or bio-simulated hyperparameters can alleviate the above problems, it is time-consuming and difficult to ensure the performance. By parametrization of the local module, the HP approach can automatically optimise the local hyperparameters and achieve a synergic learning mode. For a demonstration, we comprehensively compared the loss curves and accuracies of a single LP network, a single GP network, a fine-tuning $LP + GP$ network and the proposed HP network in Fig. 3. For fairness, we used the same initial weight configurations (see *Methods*).

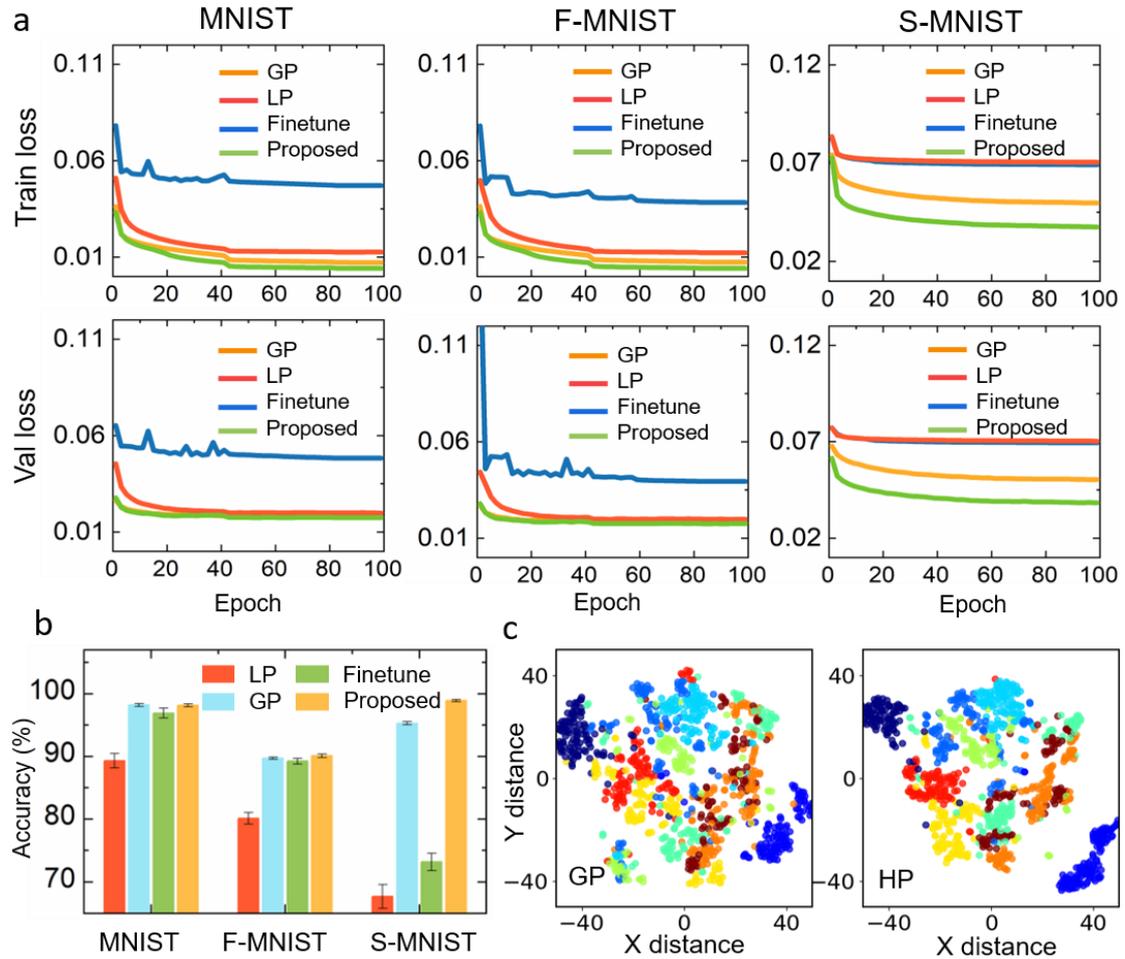

**Figure 3. The hybrid plasticity approach can effectively ensure the convergence and accuracy of hybrid learning**. **a,** Comparison of the convergence curves of the single GP-based, LP-based, fine-tuning-based (denoted by Finetune) and proposed HP model over MNIST, Fashion-MNIST (F-MNIST) and Sequential MNIST (S-MNIST). b, Accuracy histogram for the four models. The error bar indicates the standard derivation for ten repeated trials. $*P < 0.05$, Two-sided t-test was applied to assess statistical significance. c, T-distributed stochastic neighbour embedding

visualisation of the average firing rate in the first hidden layer for 1000 random test samples on S-MNIST.

Fig. 3a shows that the fine-tuning method has a poor convergence and performs worse than the other models. Through the proposed compatible design, the HP model significantly improves the accuracy of the LP model (Fig. 3b) and fine-tuning hybrid model, indicating that the proposed method can efficiently integrate LP and GP methods. Single GP model is well suitable for optimizing the errors of common classification tasks. HP model inherits the advantage and achieves a comparable convergence on static MNIST and Fashion MNIST dataset. Besides that, HP model demonstrates a higher accuracy and faster convergence than that of GP model on sequential MNIST dataset. Because the LP module provides a correlation-driven weight matrix that stores the correlation product of the presynaptic inputs and postsynaptic activations (i.e., spikes) in the previous steps, as indicated in ref. [21], it can act as a kind of attention mechanism to recent past with the strength being determined by scalar product between current hidden vector and earlier input stimulus. Combining the results of Table 1 and Fig.3, it shows that the adding of LP modules could be beneficial for the use of longer learning time-scales. Furthermore, we visualised the activations in the first hidden layer by 2D embedding visualization of T-distributed stochastic neighbour embedding over the Sequential MNIST. Take the yellow cluster in Fig.3c as an example. The adding of LP modules can help HP modules abstract points within each class more compactly and push different classes farther. Overall, the above results indicate that the HP can efficiently coordinate GP and LP methods with a stable convergence for common classification tasks.

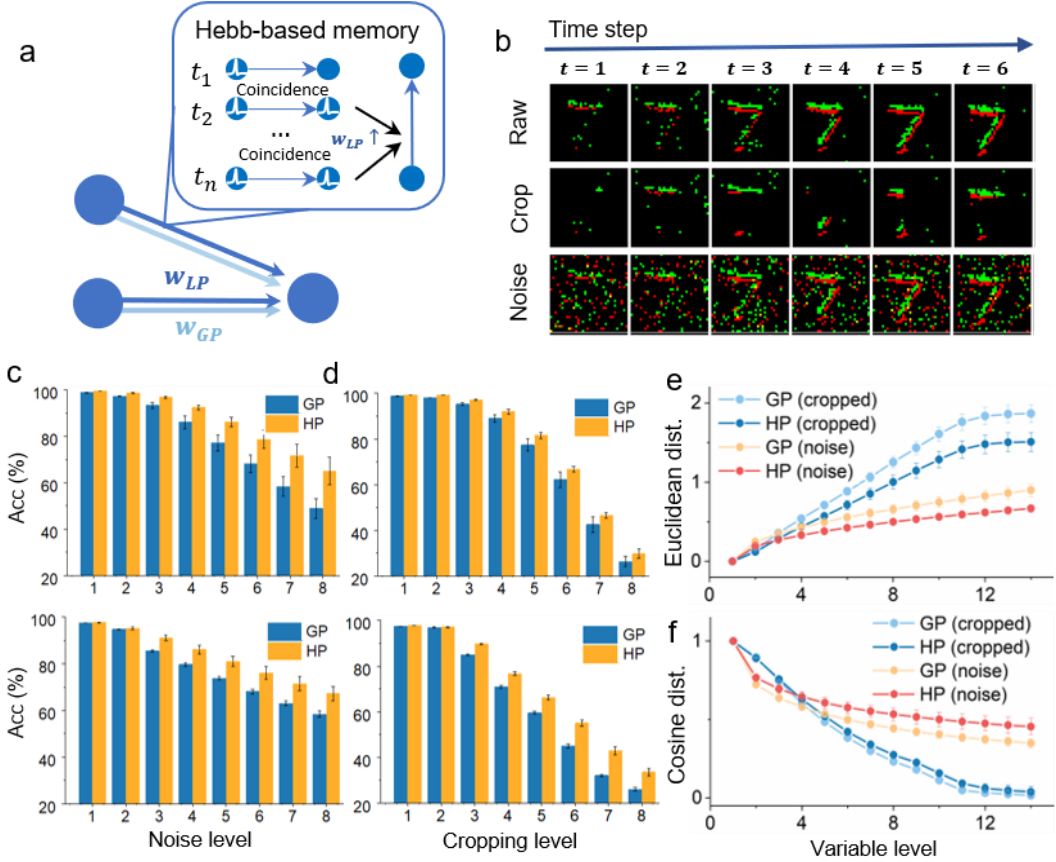

**Figure 4. Hybrid plasticity improves the learning tolerance. a**, An illustration of the memory function provided by the correlation-based LP module. When detecting a strong correlation between pre-synaptic and post-synaptic neurons, the LP module consolidates these synaptic weights with a learnable writing and decaying mechanism, which can be helpful for memorizing repeated patterns and facilitating the recognition of the similar stimulus. **b**, Generation of incomplete data by cropping (the second row) and noise mixing (the third row) using N-MNIST data. We randomly cropped raw data (the first row) with different cropping sizes, or added different levels of salt-and-pepper noise. **c**, Performance comparison in the cropping experiments using the N-MNIST (upper) and MNIST (lower). **d.** Performance comparison in the noise experiments using the N-MNIST (upper) and MNIST (lower). **e,** Euclidean distance between the hidden activation (membrane potential in the last time step) of cropped MNIST data and those of the original MNIST data. **f,** Cosine distance between the hidden activation of cropped MNIST data and original MNIST data. We plotted the average and standard deviations of ten repeated trials in c-f with *$P < 0.05$.

## 2.2 Development of the HP model for multiple learning scenarios

To demonstrate the applicability of HP model, we firstly conducted three experiments with different learning scenarios, involving the fault tolerance learning, learning with few-shot data and continual learning. We further analyse the effectiveness of the hybrid model from overall loss functions in the next section.

### 2.2.1 Improving fault-tolerance learning

Fault tolerance is important for neuromorphic chips and neuromorphic sensors to process real-time information and prevent the influence of internal noise or external interference. For example, neuromorphic vision sensors (NVS)[53] can quickly capture the per-pixel brightness changes with low latency and high dynamic range, but suffer from the inherent noise of physical devices and the movements of external background objects, thereby affecting its practical performance. Next, we demonstrate that by virtue of Hebbian-based local modules (Fig. 4a) and the HP approach, the hybrid model may improve the robustness of GP-based networks. We examined the ability of this model to handle incomplete data using an image classification dataset (MNIST) and a neuromorphic dataset (N-MNIST). We used incomplete data to refer to cropping data (e.g., parts of image information are masked, Fig. 4b) and noise-mixed data (data mixed with the salt-and-pepper noise, Fig. 4b). The models were trained on the standard datasets and tested using the incomplete test samples (see *Methods*).

Fig. 4 shows as the cropping area increases, the HP model exhibits stronger resistance to the cropping area on the N-MNIST (upper) and MNIST datasets (lower). Meanwhile, the noise experiments also show that the HP model achieves good robustness and mitigates the interference of different types of noise. Similarly, the superiority of the HP model becomes apparent as the noise level increases (Fig. 4d). To obtain a more insightful analysis, we calculated the Euclidean distance (Fig. 4e) and the cosine similarity (Fig. 4f) between the first hidden layer activation of the incomplete data and those of the original data using the same model on the MNIST dataset. As Fig.4f and Fig.4e illustrate, the HP model diminishes the pattern distance between the incomplete patterns and original patterns, which indicates that the local modules can help the network leverage the previous associative features from incomplete data and therefore benefits network fault tolerance capability.

### 2.2.2 Learning with few samples

One of hallmarks of high-level intelligence is the flexibility to learn with a severely limited number of samples. A typical machine learning scenario is few-shot learning.

In this case, the classifier must adapt to new classes not seen in the training phase when only given a limited number of samples in each class. To efficiently establish a mapping relationship from the limited data, it is vital to leverage prior knowledge or acquire inductive biases. The GP-based networks succeed in abstracting useful features; however, it is difficult for the networks to exploit prior knowledge hidden in the limited datasets without resorting to other techniques. In contrast, the human brain is highly efficient in learning from limited data. Neuroscience findings[4] reveals that the response of cortical neurons to sensory stimulus can be reliably increased after just a few repetitions by virtue of synaptic plasticity, which indicates that brain-inspired synaptic plasticity may play an important role in exploiting the correlation hidden in the limited data. Therefore, by integrating LP and GP learning, we expect that the hybrid model can solve this problem through a two-fold mechanism: (1) abstract sufficiently discriminant representation of input data mainly by the GP module; (2) find a useful inductive bias from the limited number of example pairs mainly using the LP module.

Here, we used the Omniglot dataset to examine the performance of proposed model. We adopted a widely-used network structure[24,47,54] to abstract features and compared their performance. We also allowed the training labels to feed into the last layer only during the presentation time so as to help the network establish an input-to-output correlation by local modules. In this manner, when a query sample is received, the local modules may provide an augmented signal[23,24] based on the correlation-based inter product of the query sample and the centres of each previously appeared samples. Fig. 5a-b depict the comparison results. A detailed experimental setting is provided in *Methods*. Because the vanilla backpropagation is hard to learn useful feature representations when given limited number of samples, single GP model is hardly learned in this task. The best accuracy of our model for five-way one-shot and twenty-way one-shot tasks is 98.7% and 94.6%, respectively, which are comparable with other state-of-art results and significantly higher than previous SNNs. Compared with the single GP-based model, the improved accuracy indicates that the local module plays a critical role in performance. We gave an effectiveness analysis from the perspective of metric learning (see *Method*). In addition, without resorting to additional techniques, the hybrid model can achieve competitive results that are comparable to the state-of-the-arts as shown in Table 2.

Table 2: Comparison of the published works with the similar network configurations on Omniglot datasets.

| Model | 5-way 1-shot Acc. (%) | 20-way 1-shot Acc. (%) |
|---|---|---|
| Human-level[55] | - | 95.5 |
| MAML[47] | 98.7 | 95.8 |
| Non-spiking plastic nets[24] | 98.3 | - |
| Non-spiking Siamese nets[54] | 97.3 | 88.2 |
| Spiking LN[56] | 83.8 | - |
| Spiking nets with GP (this work) | 28.4 | 8.5 |
| Spiking nets with HP (this work) | 98.7 | 94.6 |

### 2.2.3 Continual learning for multiple tasks

A pivotal point for handling multiple tasks is the capacity for continual learning, that is, an ability to learn new tasks without forgetting the previous tasks[6]. Recent studies[57] have shown that the motor cortex disinhibits a sparse subset of dendritic branches for new tasks and thereby reduces the disruption of network memory for previous tasks. This indicates that the brain may multiplex some neuro-circuits while highly activating some synaptic connections to represent task-related information for solving new tasks. These motivate us to explore a distributed hybrid learning paradigm, that is, activating a sparse overlapping subset of weight connections by GP learning and controlling other synaptic connections by a task-sharing meta-LP module. Unlike the previous XdG method[58] that uses a sub-network to solve a sub-task by masking parts of neurons in each task, our method uses a finer-grained synaptic modulation and a different synergic learning scheme. Here we allow the hybrid network to use a small number of GP-based connections to represent task-specific information, and LP learning to learn common features among tasks. By doing so, we expect to alleviate the disruption of network memory in different tasks and expand the learning capacity of hybrid networks to handle multiple tasks. In addition, because this paradigm allows the hybrid model to flexibly allocate different learning methods on different connections, this flexibility can leverage the many-core architecture to optimize the deployment of on-chip resources, such as the intra-core route and throughput, and thereby facilitates implementations on neuromorphic hardware.

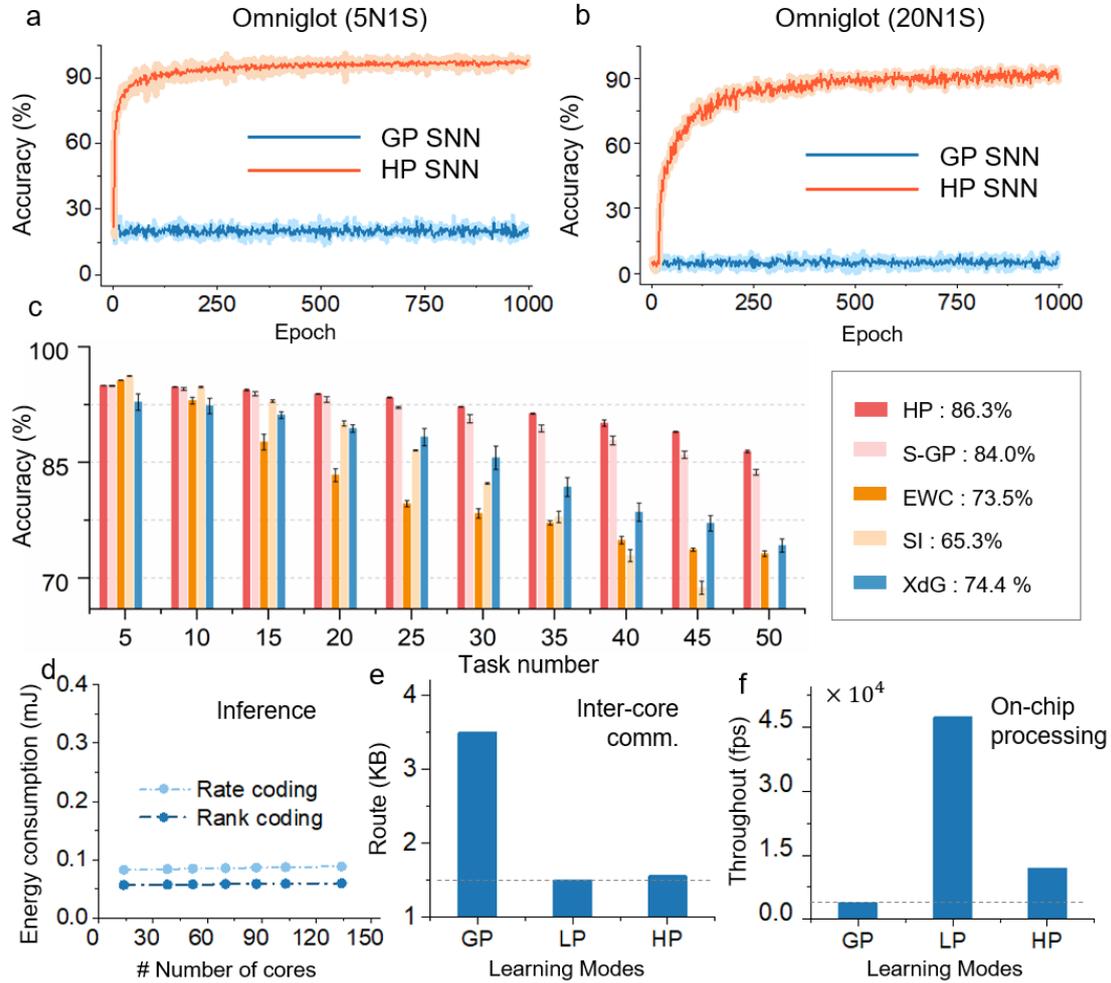

**Figure 5. Performance evaluations of hybrid plasticity spiking neural networks**. **a-b**, Performance curves on the Omniglot dataset with five-way, one-shot (5N1S, a) and twenty-way, one-shot (20N1S, b) experiments. The average and standard deviation over five runs for each epoch are reported. **c**, Histogram of the accuracy of different techniques using the shuffle-MNIST dataset. The right side shows the average performance over fifty tasks. The sparsification-weight learning is called by the S-GP for short. The error bar is given by the standard deviation over five runs. Note that all models are based on spiking leaky integrate-and-fire neurons and slightly differ from that in other published work. **d**, The HP model can support multiple spike coding schemes on the Tianjic and achieve highly-efficient inference. Critically, the energy consumption increases slowly as the network size expands owning to the spike-based paradigm and local-memory structure. **e**. The comparison of inter-core communication resources in three different learning modes. **f**. The throughput evaluation of on-chip learning in three different learning modes. Details of hardware evaluation methods are described in *Methods*.

We first examined the HP model performance on the standard shuffled MNIST dataset and compared it with single GP model and the state-of-the-art results[58-60]. We ran all models for five times and reported the testing results after fifty-task learning (Fig. 5c). We randomly activated 3% sparse and overlapping connections with GP-based learning for each task and used the LP learning to learn other connections. The meta-parameters of LP learning were trained using the first thirty-five tasks and fixed in the last fifteen tasks for evaluation. More detailed setting can be found in the *Methods*. Fig. 5c indicates that during the fifty-task learning, the HP SNNs consistently achieve the best results compared with other works. The proposed model obviously outperforms the sparse GP model, which indicates the effectiveness of proposed hybrid learning paradigm.

**2.3 Effectiveness analyses of hybrid learning model**

We next analyse the effectiveness of hybrid learning model. Because the learning of the hybrid model is affected by both the external supervision error and internal synaptic behaviours, according to different learning circuits, we assume that the overall loss of the hybrid model can be decomposed of an explicit classification loss and an implicit loss driven by the network dynamics (see *Methods*). Then we remodel the local weight update from the perspective of the optimization, and analyse the effectiveness from the approximate regularization based on hetero-associative memory (HAM)[61,62] and metric learning.

For the fault tolerance test, if we consider the local weight increment as a derivative of the implicit loss function, we can integrate the local weight increment and derived the implicit loss in the similar form of energy function used in HAM[61,62]. Similar to HAM, the Hebbian-based operation can help to encode the previous patterns triggering neuron concurrent firing behaviors into a local minimum in the energy landscape (equation (17)). On the one hand, GP learning ensures that the network can selectively activate parts of neurons firing and realize the correct response to input patterns. It indicates that the neuron concurrent firing behaviours are more likely to represent an effective response for the previous training patterns. On the other hand, the LP modules can gradually decrease the energy surface at every update based on the concurrent firing behaviours (equation (15)). Through implicitly optimizing this surface, LP modules place an approximate regularization on the network structures. It encourages the network to selectively strengthen the weights triggering these concurrent firing activities and thereby produce a stronger stimulus for the repeated or similar patterns. By combining LP and GP learning, the HP model can optimize the energy regularization and gradually relax the hierarchical representation of networks to local minimum states which may encode the previous associative patterns, thereby exploiting the correlation embedded in the appeared training examples (see *Methods*). We deliberate from the perspective of metric learning to discuss the model performance on few-shot learning. By clamping the label signals into the local module, a correlation-based LP module of the high-level input features and training labels can be established and a constraint can be placed with respect to the distribution of classes in the metric space. We prove that the LP modules can project an input pattern into a cosine-based embedding space and further produce a simple inductive bias by measuring the similarity between the query sample and the centres of each previously appeared samples (see *Method*). By doing so, the network is forced to learn from the embedding space representations to make the distance between samples within a class sufficiently small while the distance between samples from different classes sufficiently large.

Through the above analysis, we demonstrated that the LP and GP modules complement each other to form the hybrid model. An interesting finding is that we apply the hybrid model to different tasks only with minor modifications to the local module. It may provide other ways for the design of loss functions. Considering that our brain prominently uses local learning to perform tasks, transferring a part of the design of loss functions to local modules is instructive and can bring benefits from at least two-folds: (1) it can reduce the number of manually-designed hyperparameters in the overall loss functions, such as converting the original regularized weighting coefficients to model the learning rates of the local module; (2) the local-based operation endows a network with on-line and low-power properties, which especially facilitates the implementation on many-core neuromorphic hardware by virtue of the fine-grained parallelism architecture (see *Discussion*).

**2.4 Hybrid computation on the Tianjic by algorithm-hardware co-design**

The spike-based communication and local computation can be fully leveraged by using neuromorphic hardware for massive parallelism and remarkable energy efficiency. By instantiating our model on the Tianjic, next we show that the proposed hybrid model is well suitable for implementing in the dedicated hardware and provides benefit for the algorithm-hardware co-design to achieve energy-efficient and intelligent systems.

We first instantiated our model on the Tianjic neuromorphic chips (see *Methods*) and evaluated the efficiency of on-chip inferences in Fig. 5d. The proposed HP model can flexibly support the rate-based and temporal-based coding schemes and further meet different on-chip inference requirements for accuracy and inference latency (Fig. 5d). As the network size increases, owning to the spike-based paradigm and local-memory structure, the energy consumption scales very slowly. More importantly, by virtue of the neuromorphic many-core architecture, our model can significantly alleviate the overall power consumption and achieve speeds that are orders of magnitude faster than general-purpose computer (see *Table S1*).

Due to the different weight update manner, the LP and GP complement each other in the computational resources, which can be leveraged by the massive parallelism of many-core hardware for on-chip learning. Since by far there is no reported solution that can simultaneously support of LP and GP learning on many-core chips, we exploited the method of the algorithm-hardware co-design and designed a hybrid on-chip learning scheme by developing an online learning mapping scheme with developed the new cycle-accurate hardware simulator and mapping schemes, thereby evaluating the computational resources of on-chip hybrid learning. Taking the multitask learning as an example, we mainly take four steps for implementations, including mapping design scheme, software tool configuration, simulating on-chip running process and data arrangement (see *Methods*). Fig. 5e-f exhibit the simulation results of the routes and throughput of implementing the LP, GP and HP model on Tianjic chips. Our model allows the flexible configurations of these two types of learning on different connections using the proposed meta-learning paradigm. Since only a small number of weight connections are used to receive task-specific supervision signals, the workloads of inter-core communications on many-core architectures can be significantly alleviated (Fig. 5e), and the LP learning can be further deployed in core-in resources by utilizing the decentralized many-core architecture[8,63]. With the highly parallel and near-memory computing architecture of neuromorphic architecture, it can efficiently realize LP learning and facilitate on-chip learning with a high throughput (Fig. 5f).

In addition, as indicated in Fig.5e-f, the forward and backward dataflow of GP module cannot make full use of the pipelined processing mechanism of many-core chips. Thus, how to efficiently optimize the implementation of GP modules in the neuromorphic chips is an important direction to further improve the overall efficiency. Alternatively, some emerging neuromorphic hardware, such as the Loihi[63], have embedded part of X86 cores in one single chip, which provides a potential candidate for implementing GP in chips in the future. The algorithm-hardware co-design is also an important and feasible direction for development of hybrid computing paradigm. For example, the main body of many-core structure can be used to perform LP learning, while the embedded microprocessors are used to perform GP learning, which may improve hybrid on-chip efficiency and promote applications of hybrid model.

## 3  Discussion

In this work we reported a spike-based hybrid learning model which provides SNNs with an efficient synergic learning capability for handling multiple learning scenarios. Guided by the hippocampal synaptic plasticity mechanism, we developed a brain-inspired meta-learning paradigm to integrate these two types of learning. Our results indicate that with small modifications of local module, the hybrid model can achieve significantly higher performance than that of single-learning model on sequential classification tasks, and three different learning scenarios including the few-shot learning and continual learning, and fault-tolerance learning. Finally, we implemented the hybrid model in the Tianjic neuromorphic platform by exploiting algorithm-hardware co-designs and demonstrated the advantages of the proposed hybrid model on neuromorphic chips.

Understanding how the interconnected neurons in the brain combine top-down modulation information and bottom-up local information to learn to solve tasks is an active research area in the fields of neuroscience and machine learning. Three-factor learning rules provide a bio-plausible learning framework to combine global supervision information and local plasticity. Several works[11,12,37] along this line have studied the SNN's training methods. The related surrogate gradient methods[11,37] used the continuous relaxation of the gradients and provided a differentiable spiking network to update weights in a fully local computation. The e-prop[12] algorithm combined the top-down supervision signals and local eligibility traces to approximate the backpropagation of signals through time. Our model can be incorporated into a general three-factor learning framework in ref. [10], but different from the previous works, we use a meta-learning method to establish the three-factor learning model for SNNs. It exploits the biological characteristics of diverse plasticity modulation behaviors and multiscale learning mechanism, and allows to meta-learn spike-based local plasticity. In particular, we note that in the biological environment, the synaptic modulating mechanisms are more complicated, influencing a wide variety of synaptic plasticity behaviors. Unlike synapses, neuromodulators exhibit radically different evolution (learning) scales and individual functions. This coupling of neuromodulation mechanism and plasticity forms an important foundation for the biological multiscale learning mechanism[42,46]. On this basis, we formulize the hyperparameters of local learning as a type of meta parameters that can be modified by the top-down supervision signals but indirectly influences the behaviors of synapse plasticity. Furthermore, by deriving the implicit loss function from the local plasticity, we prove that the LP modules can act as a regularization over the network topology and temporal dynamics, which indicates that the roles of LP and GP modules are different from the perspective of optimization. We deliberate from the associative memory and metric learning as an illustration and further demonstrate that by constructing different correlation-based modules, the hybrid model can be linked with several existing learning models, thereby providing supports for the effectiveness analysis.

Some meta-learning-based hybrid methods have been developed to build non-spiking networks[22-26]. In the early 1990s, ref.[22] firstly proposed to optimize the parameterized local learning rule by global supervision signals. On this basis, several recent work[24,25] further established large-scale non-spiking hybrid models in the FSL task and reinforcement learning. We adopted the basic idea of learnable local plasticity as in refs.[22,24,25]. The main difference is that we use the bi-level optimization methods to decouple LP and GP learning with greater flexibilities to demonstrate the new advantages of hybrid learning in some learning scenarios, such as the robustness in fault-tolerance learning and the distributed learning capability for handling multiple tasks. Besides that, we incorporate various spiking dynamics and exploit algorithm-and-hardware co-design for prompting neuromorphic applications and hardware implementations. The bi-level optimization technique we adopted has been successfully applied in a variety of areas, such as the model transferring[64], network architecture searching[48] and has demonstrated the superiorities on solving the collaborative optimization problems. Motivated by these works[48,64], we transformed the hybrid learning into the bi-level optimization through jointly deriving from membrane potential dynamics and ion-channel dynamics and designing the hybrid learning circuits based on the special three-factor learning rule. Although several meta-learning methods[52,56,65] have recently been proposed to establish SNNs, most of studies are based on a single GP learning while lack effective mechanisms to solve the learning tasks presented in this work. By using the proposed meta-learning method and leveraging complementary hybrid learning features, our model significantly improves SNN's performance in few-shot learning tasks and further reveals several advantages of such synergic learning in multiple learning scenarios, which greatly extends the SNN's learning capability.

A salient feature of this work is the formulation of spike-based neuromorphic model. In principle, SNNs memorize historical temporal information via intrinsic neuronal dynamics and encodes information into spike trains, thereby enabling event-driven energy-efficiency computation. They are suitable for scenarios with rich spatio-temporal event-driven information and sparse dataflows, and have powered many applications in neuromorphic sensors and neuromorphic chips[51,52]. Our model retains many prominent biological attributes of spiking neurons and provides a general method to meta-learn spike-based local plasticity (see another example of meta-learning STDP rule in the *Supplementary Note*) and establish spike-based synergic learning model. By introducing complementary learning features, it may bring opportunities for current neuromorphic applications. For example, the strong robustness presented by HP models is promising to improve information processing on the neuromorphic vision sensors. Besides that, since the asynchronous spike-based communication and high-parallel computation of local plasticity can be leveraged by the neuromorphic chips, the proposed hybrid model provides a valuable opportunity to promote the efficient hardware implementation and facilitate the exploration of hybrid computing paradigm in neuromorphic hardware platform. We were aware that several recent works have reported on-line functional-level simulation schemes of meta-learning models on Loihi chips[65]. However, because the functional-level simulation simplifies the model performance environment and loses many underlying fine-grained execution details related to hardware environment, it is difficult to accurately evaluate the actual consumption and practical advantages of the hybrid learning on chips. To this end, we implemented the hybrid model in the Tianjic by developing the new cycle-accurate hardware simulator and mapping scheme. To the best of our knowledge, it is the first work to present such cycle-accurate implementation schemes on neuromorphic chips. It can prompt algorithm-hardware co-design and the exploration of hybrid computing paradigms on neuromorphic hardware architecture.

In summary, the neuromorphic hybrid learning model developed in this work exhibits a superior learning ability for solving multiple different learning tasks, and excellent energy-efficiency of hybrid computation paradigm on neuromorphic chips, which may open a new avenue of collaborative development of neuromorphic algorithms and neuromorphic computing chips.

## 5    Methods

### 5.1    Model establishment

Hybrid plastic approach is based on two sets of differential equations. The first set describes the membrane potential dynamics as follows,

$$\tau_u \frac{du_i}{dt} = -u_i(t) + \sum_{j=1}^{l_n} w_{ij}(t)s_j(t), \tag{2}$$

$$s_j(t) = \sum_{t_j^f < t} \delta(t - t_j^f), \tag{3}$$

where $w_{ij}$ denotes the weight of the synapse connecting pre-neuron $j$ and post-neuron $i$, $u_i$ denotes the membrane potential of neuron $i$, $\tau_u$ denotes the membrane time constant, $s_j(t)$ denotes the afferent spike trains, $t_j^f$ denotes the firing time, and $l_n$ denotes the number of neurons in the $l_{th}$ layer.

The second set of equations establishes on a type of diffusion dynamics of ion channels[66], modelled by

$$\tau_w \frac{dw_{ij}}{dt} = w_{ij}^g - w_{ij}(t) + P(t, pre_j(t), post_i(t); \boldsymbol{\theta}), \tag{4}$$

where $\tau_w$ denotes the synaptic constant. The first term $w_{ij}^g - w_{ij}(t)$ in the right of equation (4) denotes the recovery of $w_{ij}(t)$ into a ground state $w_{ij}^g$, which we set zero in the experiments. The second term $P(*)$ represents a general local plasticity controlled by presynaptic spike activity, $pre_j(t) \triangleq \{s_j(t)\}$, postsynaptic spike activity, $post_i(t) \triangleq \{u_i(t), s_i(t)\}$, and a group of layer-sharing controllable factors $\boldsymbol{\theta}$ which include the local learning rate, sliding threshold and other hyper-parameters and are determined by the specific local learning rules (see below).

As $P(*)$ is generic for modelling local learning rule, here we take a specific expression, a variant of Hebbian rule in our experiments, which is formulized by

$$P \triangleq k^{corr} s_j(t)(\rho(u_i(t)) + \beta_i), \tag{5}$$

where $k^{corr}$ is a weight hyper-parameter, $\rho(x)$ is a bounded nonlinear function, and $\beta_i \leq 0$ is an optional sliding threshold to control weight change directions and prevent weight explosions. It therefore can update the weight according to concurrent presynaptic firing and postsynaptic membrane activity. Integrating equation (4), we get

$$w_{ij}(t) = w_{ij}(t_{n0})e^{-\frac{t-t_{n0}}{\tau_w}} + \int_{t_{n0}}^{t} P(x, pre_j(x), post_i(x); \boldsymbol{\theta}) \, e^{-\frac{t-x}{\tau_w}} dx, \quad (6)$$

where $w_{ij}(t_{n0})$ denotes the instantaneous phasic state of synaptic weight at the phase time $t_{n0}$. Because the HP approach uses a potential trajectory rather than an equilibrium state for computation, the dependence on the initial parameter $w_{ij}(t_{n0})$ is non-trivial. Based on it, we assume that HP approach can perform supervised learning through modifying the weight phasic values $w_{ij}(t_{n0})$ at certain time $t_{n0}$, in a form of instantaneous top-down modulated signals. Consequently, we substitute the synaptic equation (6) into the membrane potential equation (2) by

$$\tau_u \frac{du_i}{dt} = -u_i(t) + \sum_{j=1}^{l_n} s_j(t) w_{ij}(t_{n0}) e^{-\frac{t-t_{n0}}{\tau_w}} + \sum_{j=1}^{l_n} \left( \int_{t_{n0}}^{t} P(x, pre_j(x), post_i(x); \boldsymbol{\theta}) \, e^{-\frac{t-x}{\tau_w}} \right) s_j(t) dx. \quad (7)$$

To enable the continuous dynamics compatible with backpropagation-based methods, we use a modified Euler method to get an explicit iterative version of equation (7)

$$\tau_u \frac{u_i(t_{m+1}) - u_i(t_m)}{dt} = -u_i(t_m) + \sum_{j=1}^{l_n} s_j(t_m) w_{ij}(t_{n0}) e^{-\frac{t_m - t_{n0}}{\tau_w}}$$
$$+ \sum_{j=1}^{l_n} s_j(t_m) \int_{t_{n0}}^{t_m} P(x, pre_j(x), post_i(x); \boldsymbol{\theta}) \, e^{-\frac{t_m - x}{\tau_w}} dx, \quad (8)$$

where we use $t_m$ to refer to the simulation timestep. Sorting the formula and substituting the specific expression of $P(x)$, we can get a set of final signal propagation equations as follows

$$\begin{cases} u_i^l(t_{m+1}) = (1 - s_i^l(t_m))(1 - k_u) u_i^l(t_m) + k_u \sum_{j=1}^{l_n} s_j^{l-1}(t_m) w_{ij}^l(t_{n0}) e^{-\frac{t_m - t_{n0}}{\tau_w}} + \\ \qquad k_u \sum_{j=1}^{l_n} s_j^{l-1}(t_m) \sum_{t_i^f, t_j^f < t_m} k_{ij}^{l,corr} H_j^{l-1}(t_m - t_j^f)(\rho(u_i(t_m))) + \beta_i) e^{-\frac{t_m - t_j^f}{\tau_w}}, \\ s_i^l(t_m) = H(u_i^l(t_m) - v_{th}), \end{cases} \quad (9)$$

where $k_u \triangleq \frac{dt}{\tau_u}$, the upper index $l$ denotes the $l_{th}$ layer, and $H(x)$ is the firing function determined by the Heaviside function. Specifically, if $u_i^l(t_m)$ exceeds the threshold $v_{th}$ $H(x) = 1$, otherwise $H(x) = 0$. Regarding the non-differentiable points of spike firing function $H(x)$, we use the surrogate gradient methods proposed by refs.[11,37] and adopt a suitable rectangle function[38] for approximating the gradient descent of SNNs. In addition, we multiply the gated signal $(1 - s_i^l)$ in the membrane attenuation term $(1 - k_u)u_i^l(t_m)$ to achieve the unique firing-and-resetting behaviour of spiking neurons.

Finally, to make the expression clearly, we replace the summation of local activity into an iterative variable $P_{ij}^l(t_m)$, relax $k_{ij}^{l,corr}$ by two elastic regular factors, $k_{ij}^{l,corr} = \alpha_i^l \eta_j^l$, and transform equation (8) into an iterative version,

$$\begin{cases} u_i^l(t_m) = (1 - s_i^l(t_{m-1}))(1 - k_u)u_i^l(t_{m-1}) + k_u \sum_{j=1}^{l_n} \left( w_{ij}^l(t_{n0}) e^{\frac{t_{n0}-t_m}{\tau_w}} + \alpha_i^l P_{ij}^l(t_m) \right) s_j^{l-1}(t_m), \\ P_{ij}^l(t_m) = P_{ij}^l(t_{m-1}) e^{-\frac{dt}{\tau_w}} + \eta_j^l s_j^{l-1}(t_m)(\rho(u_i^l) + \beta_i^l), \\ s_i^l(t_m) = H(u_i^l(t_m) - v_{th}), \end{cases} \quad (10)$$

where $dt$ denotes the length of timestep, $\alpha^l$ controls the impact of local modules and $\eta_j^l$ controls the local learning rate. It therefore formulizes the meta modules $\boldsymbol{\theta}^l$ as a group of layer-wise parameters $\{\boldsymbol{\alpha}^l, \boldsymbol{\eta}^l, \boldsymbol{\beta}^l\}$.

For classification output, we take the one-hot encoding and use $N$ neurons of the output layer to represent classification results. Then we incorporate different spike coding schemes into a general framework and describe the classification loss function $C$ by

$$C(\boldsymbol{w}, \boldsymbol{\theta}) \triangleq C\left(\boldsymbol{y}, \sum_{m=1}^T z_{t_m} q(\boldsymbol{u}^{n_l}(t_m))\right), \quad (11)$$

where $\boldsymbol{y}$ is the ground truth, $n_l$ is the number of layers, $T$ denotes the simulation windows, $C$ is a common classification loss (such as the mean square error), $q(x)$ is a non-decreasing bounded function depending on specific coding schemes, $z_{t_m} \in R_{\geq 0}$ is the weight associated with time-step $t_m$. This formulization can adapt to the rate-based coding when $z_{t_m} = 1/T$, $q(x) = H(x - v_{th})$, and adapt to the rank-based coding when $z_{t_m} = \mathbf{1}[t_m = T]$.

Given the signal propagation equations (10) and a specific decoding scheme, ideally, we can search the optimal value of network parameters, $\boldsymbol{\theta}$ and $\boldsymbol{w}$ using the BPTT algorithm. We note that the parameter $\boldsymbol{\theta}$ is a higher-level variable to modulate weights. It motivates us to exclusively establish the optimization of $\boldsymbol{\theta}$ and formulize a general expression as follows

$$\min_{\boldsymbol{\theta}} \tilde{C} \triangleq \sum_{\pi_i \in \Gamma} C_{\pi_i}^{val}(\boldsymbol{w}^*(\boldsymbol{\theta}), \boldsymbol{\theta}), \quad \text{s.t.}, \boldsymbol{w}^*(\boldsymbol{\theta}) = \arg\min_{\boldsymbol{w}} C_{\pi_i}^{train}(\boldsymbol{w}^*, \boldsymbol{\theta}), \quad (12)$$

The task $\pi_i \triangleq \{C(\boldsymbol{y}, \boldsymbol{x}, \boldsymbol{w}|\boldsymbol{\theta})\}$ samples from task distribution set $\Gamma$, and consists of a certain loss function $C$ and a set of training data and validation data. Here we used $C_{\pi_i}^{val}$ and $C_{\pi_i}^{train}$ to distinguish the loss function of training data and validation data in few-shot learning and multi-task learning. In equation (12), obtaining precision solutions of $\boldsymbol{w}^*$ is usually prohibitive and computationally expensive. To ensure the convergence and obtain feasible solutions, we formulize the above problem as a bi-level optimization that is usually used for optimizing two associated upper-level and lower-level variables[40,47,48,67]. To this end, we followed the work[47, 48] to approximate the $w^*$ by one-step gradient update in one training batch. After updating the weight $w$, we alternatively update $\theta$ across a validation task batch using the updated weights to learn fast adaptation of local plasticity. In this manner, the whole optimization can be divided into two parts by iteratively optimizing parameters $w$ and $\theta$. More concretely, in the $k$ iteration, we sampled a training task batch and updated the weights $w_k$ by the BPTT. Then we sample a validation task batch using gradient updates $w_k$ and optimize the $\theta$ over task batch by the following equation (13) until the training converges.

$$\nabla_{\theta_k} \tilde{C} = \sum_{\pi_i \in \Gamma_s} \nabla_{\theta_k} C^{val}_{\pi_i}(w^*, \theta_k) \approx \sum_{\pi_i \in \Gamma_s} \nabla_\theta C^{val}_{\pi_i}(w_k - \xi \nabla_w C^{train}_{\pi_i}(w_{k-1}, \theta_{k-1}), \theta_k), \quad (13)$$

where $\Gamma_s$ is a task batch set sampled from the task distribution set $\Gamma$, $\xi$ is the learning rate of approximation weight updates which can be helpful for accelerating the convergence[68].

### 5.2 Support rank ordering coding

By deriving from a type of ion channel dynamics model, our model maintains the synaptic decay dynamics, $k(t) = e^{\frac{t_n - t}{\tau_w}}$, during information transmission. Because the presynaptic spike signals must be filtered by the temporal filtering $k(t)$ to the post-synaptic neurons, it implies that the arriving time affects the information transition in our model. On this basis, we found a potential relationship between the HP model and the classic rank order coding assumption (see the *Supplemental Note*), and thereby developed an evidence-accumulation temporal decoding scheme. Formally, as long as the first spike is triggered by the winning neuron in the output layer, the HP SNNs will stop signal inference and produce results based on the index of the winning neuron. Then, the scaled membrane potential of the output neurons is used as an output representation to calculate the loss. In this manner, in addition to supporting conventional rate-base decoding scheme, we further utilize the synaptic dynamics and threshold mechanism of spiking neurons to implement an event-driven inference mode.

### 5.3 Details of effectiveness analyses

We further consider the impact of local module as a form of implicit loss and analyse the effectiveness from the perspective of optimization. Because the learning of the hybrid model is affected by the supervision signals and internal dynamics, accordingly, its overall loss function $E$ is more likely not only incorporating the explicit classification loss $C$, but also building on an implicit loss function $E_{in}$ generated by the inherent dynamics of network. According to the different learning circuits, we first make the decomposability assumption on the general overall loss as follows

$$E \triangleq C(t, x, y; \lambda_1 \mathbf{W}_{GP,t}, \lambda_2 \mathbf{W}_{LP,t}, \boldsymbol{\theta}) + \lambda_3 E_{in}\left(t, \{pre^l, post^l, \mathbf{w}^l, \boldsymbol{\theta}^l\}_{l=2}^{n_l}\right), \quad (14)$$

where $x$ and $y$ is external input data, $n_l$ denotes the total number of layers, and $\lambda_{1,2,3} \in R_{\geq 0}$ denote the influential factor of each part. We follow the notations of equation (1) and in a slight abuse of notation, we explicitly express the composition $\lambda_1 \mathbf{W}_{GP,t}, \lambda_2 \mathbf{W}_{LP,t}$ on $C$ to highlight the difference between the hybrid model and single-learning based model. Indeed, the expression of $E$ can be regarded as an extension of single learning-based network. In the case of $\lambda_2 = \lambda_3 = 0$, $E$ degenerates to the conventional classification loss for the GP-based network, and in the case of $\lambda_1 = 0$, the network reduces to the LP-based network.

### 5.3.1 Approximate regularization

If we treat the local weight increment as an implicit derivative for a part of the overall loss function $E$, it inspires us to integrate its weight increment to obtain the implicit loss function $E_{in}$. On the other hand, not all local weight rules can easily derive an integral function. For simplicity, we mainly focus on the impact of local weight rules on the optimization of the current layer weight $w$ and illustrate the effectiveness of local modules for the specific task.

In the fault tolerance learning, we accordingly treat the local weight increment $\Delta w_{ij,t}^l$ as the implicit derivative of $E_{in}^l$ by

$$\frac{\partial E_{in}}{\partial w_{ij}^l} \propto -\Delta w_{ij,t}^l \propto -s_{j,t}^{l-1} \rho(u_{i,t}^l). \quad (15)$$

We use Hebbian rules and set $\rho(u_{i,t}^l) = H(u_{i,t}^l - v_{th})$ in the derivation. Integrating above equation, we can get a loss expression $E_{in}$ as follows,

$$E_{in} \approx -\sum_{t=1}^{T}\sum_{l=2}^{n_l} \mathbf{s}_t^{l-1^T} \int H(\mathbf{u}_t^l - v_{th}) d\mathbf{w}^l = -\sum_{t=1}^{T}\sum_{l=2}^{n_l} \mathbf{s}_t^{l-1^T}\left(\mathbf{w}_t^l \mathbf{s}_t^l - \int \mathbf{w}^l \frac{dH}{d\mathbf{w}^l}\right), \quad (16)$$

Since the derivative of the Heaviside function is clearly zero for $u \neq v_{th}$, the following equation holds

$$E_{in} \approx -\sum_{t=1}^{T}\sum_{l=2}^{n_l} \mathbf{s}_t^{l-1^T} \mathbf{w}_t^l \mathbf{s}_t^l, (u \neq v_{th}). \quad (17)$$

We note that a form of loss function in equation (17) is similar to the energy function used in hetero-associative memory (HAM)[61,62,69] in which Hebbian-based operations help network encode the previous associative patterns into the local minimum of the energy surface. It inspires us to explain the model effectiveness from the optimization of the similar energy function. Specifically, in the HP models, the supervised GP learning ensures that the network can selectively activate parts of neurons firing and realizes a correct response to input patterns. Thus, the associative patterns of neuron concurrent firing behaviours (i.e., $s_t^{l-1^T} s_t^l$) are more likely to represent an effective response for input patterns. On the other hand, as shown in equation (17), Hebbian-based operation can decrease the surface at every update. Therefore, by optimizing the energy surface, the LP module places an approximate regularization on the network structures with the punishment of $-\sum_{l=2}^{n_l} s_t^{l-1^T} w_t^l s_t^l$. It encourages the network to strengthen the weights triggering neuron concurrent firing behaviours, resulting in a stronger stimulus for similar or repeated patterns. Collectively, by combing the LP and GP modules, the HP model can relax the hierarchical representation of the networks to local minimum states that are more likely to encode an effective response to the previous associative patterns, thereby exploiting the correlation embedded in the appeared training patterns for the recognition of incomplete patterns.

Please note that unlike HAM models using one or more bi-directional iterations for pattern reconstruction[61,62], the HP model leverages the memory matrix for the classification of disturbance patterns. Hence, we provide another intuitive explanation to analyse the correlation-based augmented information provided by the LP modules. Given a general input and output dataset $D = \{(x_i, y_i)\}_i^N$, where $y_i \in R^{m \times 1}$ refers to the response of the current layer to the input $x_i$, let us assume that a querying sample $\tilde{x} \in R^{n \times 1}$ is received and belongs to the $D_k$ category. Then the local module produces an augmented information $I_{LP}$ by

$$I_{LP} = w_{LP}\tilde{x} = \sum_j y_j x_j^T \tilde{x} = \sum_{x_k \in D_k} y_k (x_k^T \tilde{x}) + \sum_{x_j \notin D_k} y_j (x_j^T \tilde{x}). \quad (18)$$

When the network receives a disturbance sample $\tilde{x}$, $w_{LP}$ provides an augmented information of the previously stored pattern $y_i$ in the form of a weighting coefficient $(x_k^T \tilde{x})$. Since the inner product from the same class can provide a stronger augmented information, it also indicates that the local module can exploit the correlation between the input sample and the previous appeared associative patterns, thereby facilitating the classification of disturbance patterns.

### 5.3.2  Adding inductive biases of prior knowledge

Next, we illustrate the correlation-based local modules can place a constraint with respect to the distribution of classes in the cosine-based metric space for accelerating the convergence of HP models. Assume that we have a set of training samples $D = \{(x_k, y_k)\}_{k=1}^{N_D}$ where $x_k \in R^{m \times 1}$ is an $m$-dimensional feature vector, $y_k \in R^{n \times 1}$ is the one-hot label, $N_D$ denotes the sample number of dataset $D$. We refer $x_k$ to the general features coming from raw data or the last $n_l - 1$ layer. By introducing the training labels to the output neurons, the local module can construct the Hebbian-like matrix as follows

$$w_{LP} = \sum_{k \in N_D} \eta_k y_k x_k^T = \sum_{k=1}^{K} (y_k \sum_{i=1}^{N_{D_k}} \eta_i x_i^T) = \sum_{k=1}^{K} y_k c_k^T, \quad (19)$$

where $K$ denotes the class number of $D$ and $D_k$ denotes the sub-set of examples within the same class. Here we set the learning rate $\eta_i = \frac{1}{N_{D_k} \|x_i\|_2}$ and refer $c_k = \frac{1}{N_{D_k}} \sum_{i=1}^{N_{D_k}} \frac{x_i}{\|x_i\|_2}$ to the sample mean of the class $D_k$. To keep the clarity of proof, we also simplify the modelling of meta-parameters, such as the gating parameter. Based on equation (18), when entering a query sample $\tilde{x}$, the local module produces an inductive bias $I_{LP}$ by $I_{LP}(x) = \sum_{k=1}^{K} y_k (c_k^T \tilde{x})$. Then the membrane potential of output neuron is governed by

$$u_i^{n_l}(t_m) = \lambda_1 I_{GP,i}(t_m) + (1 - \lambda_1) I_{LP,i}(t_m) = \lambda_1 \sum_{j=1}^{l_n - 1} w_{ij} \tilde{x}_j(t_m) + (1 - \lambda_1) \sum_{k=1}^{K} y_{k,i} c_k \tilde{x}_j(t_m), (20)$$

where $y_{k,i}$ is the $i$ element of the one-hot label $y_k$, which satisfies that $y_{k,i} = \mathbf{1}_{\{k\}}(i)$. In the experiment, we initialize $\lambda_1$ with a small positive value to strengthen the impact of $I_{LP}$ in the early training phase. In this manner, we can give an intuitive interpretation for the inductive bias $I_{LP}$ from the Euclidean distance $L$ between the label $y$ and $u^{l_n}$. Assuming that the input pattern belongs to $i$ class, $L$ can be calculated by

$$L(u^{l_n}, y) = \sum_{q \neq i} (\lambda_1 I_{GP,q} + (1 - \lambda_1)(c_q^T \tilde{x}))^2 + \left(1 - ((\lambda_1 I_{GP,i} + (1 - \lambda_1)(c_i^T \tilde{x}))\right)^2 =$$
$$\sum_{q \neq i} (\lambda_1 I_{GP,q} + (1 - \lambda_1) c_q^T \tilde{x})^2 + (1 - c_i^T \tilde{x} - \lambda_1((I_{GP,i} - c_i^T \tilde{x}))^2, \quad (21)$$

where we set $T = 1$ and omit the index $t$ for neatness. Note that the $\lambda_1$ is a pre-defined small amount, thus, the $L_\phi$ has the two main components $(1 - c_i^T \tilde{x})^2$ and $(1 - \lambda_1)^2 c_q^T \tilde{x}$. To minimize the distance $L$, the network is forced to learn the specific feature mapping that projects the distance between samples within a class sufficiently small (by the punishment of $(1 - c_i^T \tilde{x})^2$) while the distance between samples from different classess sufficiently large (by the punishment of $c_q^T \tilde{x}$). In this manner, the inductive bias $I_{LP}$ enables the HP model to learn from the feature similarity between the query data and the previous feature centres in the training data in the metric space.

## 5.4 Task-related details

### 5.4.1 Basic performance evaluation

In the MNIST and Fashion-MNIST experiments, we used Bernoulli sampling to encode the pixel into spike data. In the sequential MNIST and CIFAR10, we used the first spiking layer as an encoding layer[38] to produce spike signals. Here the sequential MNIST is a variant of MNIST datasets, which inputs the original image in a row-by-row pixel manner. In the CIFAR10-DVS and DVS-gesture, we accumulated spike trains (8ms and 10ms, respectively) for acceleration and directly input them into SNNs. We applied the batch normalization (BN) technique to convolutional layers on the DVS-Gesture dataset by following the work[35]. We optimized the HP model in all datasets using the mean square error (MSE) and the adaptive moment estimation optimizer (ADMM). A four-layer MLP with [28-256FC-256FC-10] was trained on sequential MNIST, a six-layer CNN with [input-128C3-AP2-256C3-AP2-256C3-AP2-512FC-10] was trained on MNIST and Fashion-MNIST datasets, a nine-layer CNN with [Input-64C3S2-BN-128C3S1-BN-256C3S1-BN-256C3S1-BN-256C3S1-BN-256C3S1-AP2-800 FC-512FC-11FC] was trained on DVS-Gesture datasets, and a nine-layer CNN with CIFARNet structure[38] was trained on CIFAR10 and CIFAR10-DVS. We took the local learning rules in equation (5) and meta-parameters in equation (10) in all experiments unless otherwise stated. In the comparison with fine-tuning models, we used the same learning rules and fixed these meta-parameters after random initialization during training. To reduce computation, we equipped the local synaptic module in the hidden fully-connected layers in all classification experiments. Other network parameter configurations can be found in the Table S2.

### 5.4.2 Fault tolerance learning

In the cropping experiment, we increased cropping area gradually on the centre of each image or each NVS frame, denoted by $(2ci)^2$, where ci represents for cropping intensity with a range of $0{\sim}14$. In the noise experiment, since salt-and-pepper noise can maintain the spike binary representation, we use it for evaluating the robustness to noise in N-MNIST and MNIST experiments. We also increased the proportion of noise region on each image or each NVS frame gradually, denotes by the noise-level (*nl*), where $nl$ value refers to the $nl \times 2e^{-2}$ region with a range of 0~14. All models were pre-trained on the standard training dataset and tested on the cropping (noise) data with the same parameter configurations, network structure [input-512FC-10FC], and the MSE loss. For the distance comparison, we calculated the membrane potential of the first hidden layer in the last timestep as the representations to calculate the distance between the incomplete data and the original data. We randomly sampled 1000 testing data from MNIST and plotted its average distance on Fig. 4e and Fig. 4f.

### 5.4.3 Few-shot learning

The Omniglot is a standard few-shot learning dataset that contains 1623 categories and each category contains 20 samples. First, we randomly selected N classes and sample S sample pairs from each class (called N-way S-shot) for one-episode training. Then, in the testing, we first sampled S labelled samples from each N testing classes (named by the presentation time) and fed into the classifier. Then, we randomly sampled a new but unlabelled instance from the same N classes and query the classifier for its label. We used 4 convolutional layers with $3 \times 3$ kernel size and 2 strides, followed by a fully connected layer and then a N-way classifier layer. We followed the work[24] to configure the network parameters and divide the training sets and testing sets. During the training process, we sampled five task episodes from the task distribution $\Gamma$, and used the one-step updated weights $w$ to approximate $w^*$ using five training task samples, and then alternatively iterated the meta-parameters $\theta$ by re-sampling validation samples from the same tasks. To reduce the computation, we equipped the local plasticity module into the fully-connected layer. The training label was fed into the last classifier layer by a one-hot coding scheme to guide the correct classifications. We used the first layer as an encoding layer that converts the real-value images into spike signals as in refs.[26, 38]. We trained the network 3000000 episodes and reported the best results over the last 1000 episodes.

### 5.4.4 Continual learning

The shuffled MNIST experiments include several sequential learning tasks. All tasks are to classify handwritten digits from zero to nine. For each new task, the image pixels were randomly permuted with the same randomization across all digits in the same task and different randomization over different tasks. We trained each task by ten epochs and used a four-layer spiking networks with [784-1024-1024-10] MLP structure to minimize the MSE by the ADMM. During the training of each task, we fixed the meta-parameters of LP modules and randomly generated a sparse and fixed connection matrix to receive supervision signals. We used GP learning to update these sparse weights, and LP learning to update other connections. After each task is learned, we fixed the weights and randomly took validation task samples from the learned task set $\Gamma$ to update LP modules for one epoch. Other comparison techniques were adopted from corresponding publication and applied to spiking models.

**5.5 Details of Hardware implementation**

Neuromorphic computing systems aim to emulate biological neural coding and computation in hardware by distributing memory and computation in a large number of self-contained computational units. The sparse spike-triggered paradigm and the colocation of memory and compute in neuromorphic systems lead to a highly efficient computing paradigm, and the decentralized many-core architecture can significantly increase the parallelism, thereby providing orders of magnitude greater efficiency than general-purpose computers. Tianjic, a unified network computing platform, is an advanced many-core neuromorphic system with innovations in hybrid computing paradigm and architecture. In the following, we give a brief overview of Tianjic neuromorphic systems and our model implementations thereon.

The Tianjic chip is a cross-paradigm neuromorphic computing platform that can support a broad spectrum of neural coding schemes, computational models and network structures. It is fully digital and fabricated using 28-nm high-performance low-power technology. Each Tianjic chip contains 156 functional cores (FCores) which are arranged in a 2D-mesh manner. Each FCore contains a group of neurons, a group of axons and synaptic connections between them. Among the FCores, spikes can be transmitted to one or more cores in the mesh through the routing network in the form of routing packets. At the same time, through the inter-chip communication interface, multiple chips can expand the internal routing network connections into a larger computing platform.

We mapped our model onto multiple FCores, where different FCores are configured to perform different basic operations and transformations. Taking the MNIST dataset as an example, we deployed 70 FCores to implement a fully-connected structure [784-1024-1024-10]. We tested its energy consumptions with different coding schemes. Because the rank coding shortens the average decision time, it can effectively reduce the on-chip inference latency and the average compute ratio, thereby alleviating average dynamic power consumption. Specifically, the average on-chip inference latency required for rate coding and rank coding are 0.27us, 0.18us, the compute ratios are 0.63, 0.45, and the dynamic power consumptions are 0.48W, 0.38W, respectively. We reported the on-chip performance on MNIST, F-MNIST and N-MNIST datasets and compared it with GPU-based running results in supplementary Table S1. With the massive parallelism and the near-memory computing architecture, the execution time on the Tianjic can be much faster than that of the general-purpose computer. Importantly, the energy consumptions scale only slightly as network size increases owning to the spike-driven paradigm and local-memory many-core architecture (Fig. 5d).

Since currently there is no reported neuromorphic chip that can simultaneously support on-chip LP and GP learning, we based on the Tianjic's hybrid structure to design an on-chip hybrid learning scheme (see the *Supplementary Note* and *Extended Figure S1*) and feasible cycle-accurate hardware simulation scheme to evaluate on-chip computational resources. Because the cycle-accurate simulator can well capture the hardware chip properties at runtime, it is commonly used for chip evaluation. Note that here we took an extended version of released Tianjic neuromorphic cross-paradigm chip with enhanced re-configurability and functionalities to support continuous execution of multiple operations. On this basis, we developed a mapping scheme to disassemble the overall dataflow into performable fine-grained basic operations, and further transformed a mapping design into executable configuration. By doing so, we can simulate three on-chip learning modes (LP, GP and HP) using the software tool chain. A detailed simulation scheme is provided in the Extended figure 1. With this simulation scheme, we estimated the throughput and route cost of different learning modes using a MLP structure [784-512-10] and the time window $T = 3$. Regarding the route cost in fig. 5e, we accumulated the amount of data volume whenever data transmission occurs. Regarding the throughput in fig. 5f, we recorded the time spent in each phase when executing computational tasks on all allocated FCores. After that, we summed the time consumptions together to count the total clock consumption, and thereby the throughput. More details of evaluation methods are showed in *Supplementary Note*.

**Competing interests**

The authors declare no competing interests.

**Data and code availability**

All data used in this paper are publicly available and can be accessed at http://yann.lecun.com/exdb/mnist/ for the MNIST dataset, https://www.cs.toronto.edu/~kriz/cifar/

for the CIFAR dataset, https://www.garrickorchard.com/datasets/n-mnist for the N-MNIST dataset, https://github.com/brendenlake/omniglot/ for the Omniglot dataset.

# 6   Method References

# 7 Supplementary information

## 7.1 Extended Data Table 1: Performance on Tianjic chips

**Supplementary Table 1.** Performance evaluation of implementations of hybrid models on different computational platforms.

| Dataset | Platform | Coding | Acc. (%) | Compute ratio | Power (W) | Energy (mJ) | Latency (ms) |
|---|---|---|---|---|---|---|---|
| MNIST | GPU | Rate | 99.55 | - | 37.13 | 338.99 | 9.13 |
|  | Tianjic | Rate | 99.43 | 0.46 | 9.35 | 3.46 | 0.37 |
|  | Tianjic | Rank | 99.22 | 0.27 | 9.22 | 2.56 | 0.23 |
| F-MNIST | GPU | Rate | 93.45 | - | 39.23 | 588.24 | 14.23 |
|  | Tianjic | Rate | 93.30 | 0.46 | 9.56 | 3.53 | 0.37 |
|  | Tianjic | Rank | 93.11 | 0.33 | 9.37 | 2.80 | 0.30 |
| N-MNIST | GPU | Rate | 99.53 | - | 49.97 | 1214.8 | 24.31 |
|  | Tianjic | Rate | 99.45 | 0.46 | 13.31 | 4.93 | 0.37 |
|  | Tianjic | Rank | 99.21 | 0.28 | 13.00 | 3.65 | 0.28 |

## 7.2 Extended Data Table 2: Parameter Configurations on different datasets

**Supplementary Table 2**: Hyper-parameter setting on different classification tasks.

| Parameters | Descriptions | MNIST | FMNIST | CIFAR10 | DVS-CIFAR10 | DVS-Gesture | Omniglot |
|---|---|---|---|---|---|---|---|
| dt | Simulation timetep | 1ms | 1ms | 1ms | 8ms | 10ms | 1ms |
| Batch size | - | 100 | 100 | 50 | 32 | 12 | 3 |
| $a$ | Gradient Width of H | 0.5 | 0.5 | 0.5 | 0.5 | 0.5 | 0.5 |
| $N_0$ | Training epochs | 100 | 100 | 150 | 150 | 150 | 1000 |
| $\tau_w$ | Equation (10) | 40ms | 40ms | 40ms | 100ms | 100ms | N/A |
| $k_u$ | Equation (10) | 0.6 | 0.7 | 0.5 | 0.3 | 0.3 | 0.2 |
| $v_{th}$ | Equation (10) | 0.3 | 0.4 | 0.5 | 0.4 | 0.4 | 0.5 |

## 7.3 Supplementary Note 1: Design of three learning methods on Tianjic chips

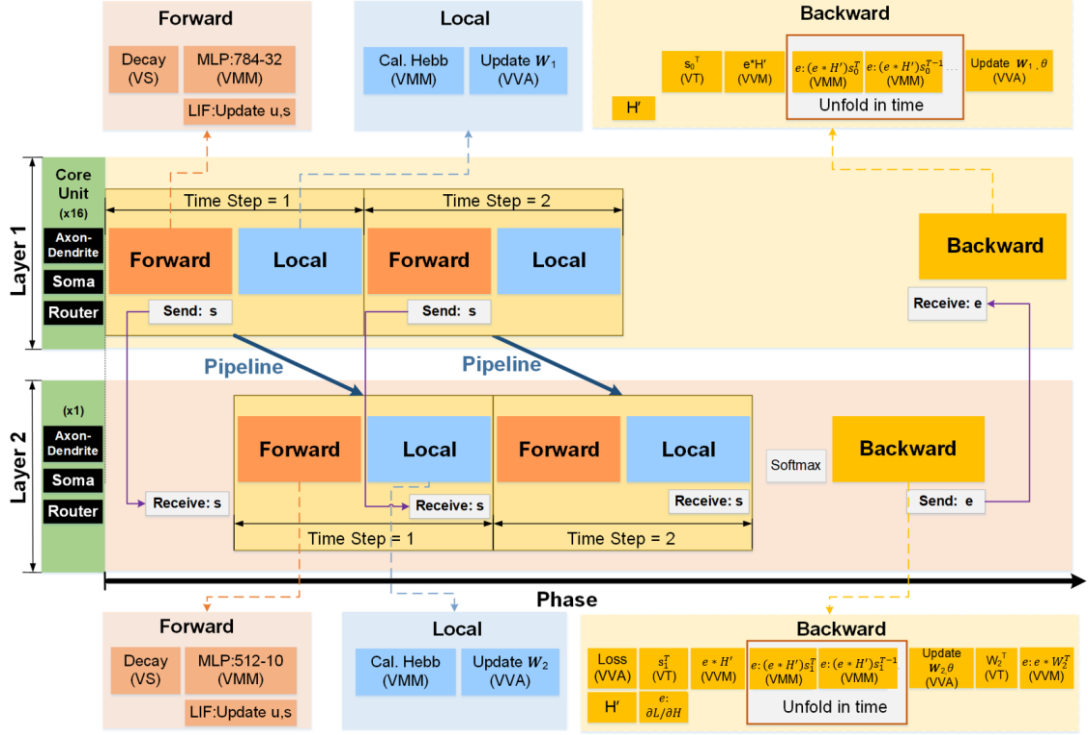

**Supplementary Figure S1.** Illustration of our hybrid on-chip learning implementation schemes. The left black boxes denote the main units in FCores, in which every operation is executed and aligned in phasic form. The inference process contains two modules: a forward module in jacinth boxes and a local module in light blue, carrying out for T times. The following backward process (the orange boxes) calculates error information and propagate in reverse direction, in order to update weights parameters and meta-parameter $\theta$. The purple arrows represent cross-layer routing paths. The forward path and local path are performed in a pipelined manner.

We designed a comprehensive on-chip hybrid learning scheme with a configurable software tool chain from algorithm to hardware to emulate the forward and backward paths in three learning modes. Specifically, we developed the dual spatial-temporal unfolding mapping scheme (called DST-UM) and the mapping compiler for network partition and resource placement, and the cycle-accurate simulator that satisfies chip-specific constraints. In this manner, we can evaluate performance under various workloads, including single-paradigm and hybrid-paradigm modes. Here we present the mapping implementation scheme of hybrid on-chip learning in Fig. S1.

As shown in Fig. S1, we disassembled the dataflow of the entire hybrid learning into basic operations which can be supported by the hardware platform, where the calculation between input spikes and weight is split into multiple spatial basic operations due to resource limitation, including fan-in and fan-out, memory space in FCores. In addition, the proposed DST-UM scheme enables the execution dataflow to be allocated and optimized in single spatial dimension or temporal dimension. For the spatial dimension, the parameters and the computation of the first layer are allocated into 16 FCores for balancing computation and memory overhead. Meanwhile the task of second layer was allocated into another FCore. On this basis, layer-wise data communication and arrangement can be performed across multiple FCores. For the temporal dimension, the DST-UM unfolded the operations of forward and backward passes along with the timing order. As shwon in Fig. S1, the forward path was organized along with time window and the backpropagation path was unfolded according to the derivation compute dependence. By doing so, the operation of the entire process can be mapped onto one chip in a spatially parallel and temporally serial manner. Finally, we organized this procedure further according to algorithmic details. In the inference process, we propagated spike signals layer-by-layer and updated local items (the orange boxes in Fig. S1) in each timestep. After performing the inference for T times, we updated the weights and meta-learning parameters $\boldsymbol{\theta}$ along the backward path (the orange blue boxes in Fig. S1).

Given above mapping implementation, we used a software tool chain reported by ref.[65], which includes a mapping compiler used for network partition and resource placement, and a C++ based cycle-accurate simulator used for hardware simulation that considers all chip-specific constraints. We used the compiler to perform placement and generate configure file automatically. Therefore, our simulator can utilize the configuration and simulate the running process to generate the corresponding analysis results and evaluation. We configured each FCore according to Fig.S1 and thereby used the software tool train to simulate the entire process of hybrid learning. Notably, we also simulate the event-driven attributes and the sparse processing of intermediate variables in three modes. We utilized the parallel processing mechanism of the many-core architecture to carry out the inference process and LP learning in a pipelined manner and performed the backward process of GP and HP in serial manner. Finally, we base on this scheme and estimate the computational resources of on-line learning.

### 7.4 Supplementary Note 2: Details of hardware evaluation methods

We chose the three models (LP, GP, and HP) with the same MLP structure [784-512-10] as an example, and set the time window at T =3. We mainly taken three steps during implementation, which are described as follows:

(1) **Designing mapping scheme**. We used the Tianjic mapping scheme[60] to disassemble the overall dataflow into performable fine-grained basic operations. Considering the huge number of parameters in the first layer, we allocated the computation task of the first layer into 16 function cores (FCores) for balancing the compute and memory overhead. Then, we allocated the second layer into one FCore, where layer-wise data communication and arrangement can be performed;

(2) **Configuring software tool chain.** Based on the mapping design, we further determined the key operations and emulated the process of on-chip learning (see *Extended Figure 1* for more detailed implementations). All operations are divided into three basic units of FCores, including Axon-Dendrite, Soma and Router. We initialized the network parameters of specific models according to hardware constraints and further transformed the mapping scheme into a specific executable configuration with the Tianjic software simulation tool chain[65]. By doing so, we can simulate the three on-chip learning modes (LP, GP and HP) using the Tianjic software platform.

(3). **Simulating on-chip running process and data arrangement**. With the proposed mapping scheme, we can estimate the entire online learning process of different learning modes using the Tianjic simulacoted that in the Tianjic, the entire process is executed via a group of phases in a sequential manner, where each phase contains the operations from the subset of three basic units (Axon-Dendrite, Soma and Router). Hence, we can collect the associated running cost of each operator in each unit, such as memory consumption, communication data volume and running clock cycles. And then we obtained the implementation cost thereon. For better illustrating the evaluation, we first briefly introduce some basic concepts and definitions, and then elaborate the specific computation methods related to the throughput and route cost.

**Mathematical abstract preliminaries of simulation process.** Assume a task process is performed as a set of phases $\mathbb{C} = \{Phase_1, ..., Phase_i, ..., Phase_N\}$, wherein $N$ denotes the number of phases required to run a complete cycle. The $i_{th}$ phase $Phase_i$ is an operator set which contains fine-grained functional operations supported by the three basic units (i.e., Axon-Dendrite, Soma and Router). If we denote the operator sets of Axon-Dendrite unit, Soma unit and Router unit by $O_{AD}, O_S, O_R$, respectively, it yields $Phase_i \subseteq \{O_{AD}, O_S, O_R\}$. Please note that in the throughput cost evaluation, we use $O_u$ to refer to any of the three operator sets, and configure $O_u$ according to the task requirements and mapping schemes to realize the calculation of the three models as showed in Extended figure 1.

**Regarding to the route cost evaluation.** The route cost (RC) yields the following equation:

$$RC = \sum_{i \in \{v | Phase_v \cap O_R \neq \emptyset\}} CoreN_i \cdot \sum_{q}^{Q_i} Packet_q.$$

Here the communication data volume $Packet_q$ is carried by the $q_{th}$ route packet and measured through simulating the running process. $Q_i$ denotes the total number of route packet in the $i_{th}$ phase. $CoreN_i$ denotes the allocated number of FCores performing one task in parallel in the $i_{th}$ phase. $Phase_i \cap O_R \neq \emptyset$ means that we only record the phases involving with inter-core communication.

**Regarding to the throughput cost evaluation.** We recorded the time spent in each phase when executing computational tasks on all allocated Cores. The throughput cost (TC) yields the following equation:

$$TC = \frac{F}{\sum_{i=1}^{N} \sum_{u \in \{v | O_v \in Phase_i\}} \max_{k \in \{1,2,..,CoreN_i\}} Clock_k^u},$$

where F denotes the clock frequency of the simulator, u denotes different operator type from $\{O_{AD}, O_S, O_R\}$, $Clock_k^u$ denotes clock consumption of the $k_{th}$ allocated core in the $u_{th}$ type operator of $Phase_i$. Here we record the max number of the $u_{th}$ operator among $CoreN_i$ cores in $Phase_i$ for computation.

On this basis, we accumulated the data volume whenever data transmission occurs, and summed the time consumptions together to count the total clock cycles. By combining the recorded data and evaluation formulas, we can compute the route cost and throughput cost, and then obtained the results in Fig.5e and Fig.5f.

**Regarding to the energy consumption.** We pretrained HP models with different network sizes and allocate the FCores for corresponding network size. On this basis, we ran the inference process on the Tianjic using developed mapping tools[11,57] and measured the power consumption and running time. Finally, we multiplied them and obtained the energy consumption shown in Fig.5d.

### 7.5 Supplementary Note 3: Linking the HP SNNs with rank order coding

We give an overview of rank order coding[50] and analyse its relationship with the HP model thereon.

***Rank order coding*** assumes that the biological neurons encode information by the firing orders across a neuron population. Suppose target neuron $i$ receives inputs from a presynaptic neuron population $A_l$, and each neuron only fires a spike once. Let activations of afferent neurons be $a_j^l$. Then the rank order coding can record the relative firing orders of afferent neurons, and updates the activation of $a_i^{l+1}$ by,

$$a_i^{l+1} = \sum_{j \in A_l} r^{order(a_j^l)} w_{ij}^{l+1}, \tag{1*}$$

where $r \in (0,1)$ is a given punishment constant, $order(a_j^l)$ is the firing order of neuron $j$ in the presynaptic population. Equation (1*) shows that the ranking factor $r^{order(a_j^l)}$ is a key for the rank order coding, which can encourage the early firing of neuron meanwhile punish the later firing of neuron. As the encoded information is prominent in earlier spikes, this coding scheme is more situable for fast-decision and network sparsity[50]. Next, we show that the ranking factor can be equivalently converted into the decay function of our model, and therefore the information propagation of HP model encodes information by equation (1*).

***Theorem 1*** *Assume each neuron fires at most one spike in a given short time window. The HP model encodes information in the form of rank order coding.*

***Proof.*** According to the assumption, we first formulate the input current I of equation (1*) as

$$I = \sum_{j=1}^{l_n} r^{order(s_j^l)} w_{ij}^{l+1} s_j^l, \tag{2*}$$

where we add a spike signal $s_j^l \in \{0,1\}$ to incorporate all presynaptic neurons and make the updating compatible with our neuron update equations. By formula deformations, it holds

$$I = \sum_{j=1}^{l_n} w_{ij}^{l+1} e^{\log(r) order(s_j^l)} s_j^l = \sum_{j=1}^{l_n} w_{ij}^{l+1} e^{-\frac{order(s_j^l)}{\frac{1}{\log(r)}}} s_j^l = \sum_{j=1}^{l_n} (w_{ij}^{l+1} e^{-\frac{t_m - t_0}{\tau_w}}) s_j^l, \tag{3*}$$

where $\tau_w = \frac{1}{-\log(r)}$, $order(s_j) = t_m - t_0$. In this manner, the ranking factor $r^{order(s_j^l)}$ can be equivalently converted into the decay function of the HP method.

### 7.6 Supplementary Note 4: Meta-learning spike timing-dependent learning (STDP) rule

HP model provides a general method to learn the spike-based local plasticity by modifying the plasticity function $P(t, pre_j(t), post_i(t), w_{ij}; \boldsymbol{\theta})$ and using the proposed meta-learning method. Here we take the STDP rule as another demonstration.

We use the synaptic trace method[28] to record the pre- and post-neuronal firing activity and established an STDP-based HP model. Specifically, we use a synaptic trace variable $x_{pre}$ to keep the spike history of each pre-synaptic neuron as follows

$$\tau_s \frac{dx_{pre}}{dt} = -x_{pre} + \Sigma_{t_f<t}\, \delta(t-t_f), \tag{4*}$$

where $\tau_s$ denotes the decay constant of synaptic trace. When a spike arrives at this synapse, $x_{pre}$ is increased by 1, otherwise $x_{pre}$ decays exponentially by a decay factor $\tau_s$. Likewise, we use $x_{post}$ to keep the spike history of each post-synaptic neuron. By doing so, the local plasticity variables $P$ can be replaced by

$$P(t_m) = P(t_{m-1})e^{-\frac{dt}{\tau_w}} + A_{pre}x_{pre} - A_{post}x_{post}, \tag{5*}$$

where the local hyper-parameter $A_{pre}$ and $A_{post}$ can be meta-learned by our proposed methods.